\documentclass{esannV2}
\usepackage[dvips]{graphicx}
\usepackage[latin1]{inputenc}
\usepackage{amssymb,amsmath,array}
\usepackage{cite}
\usepackage{gensymb}
\usepackage{caption}
\usepackage{pdfpages}

\begin{document}
\title{Don't waste SAM}
\author{Nermeen Abou Baker$^1$ 	and Uwe Handmann$^1$
\thanks{This work has been funded by the Ministry of Economy, Innovation, Digitization, and Energy of the State of North Rhine-Westphalia within the project Prosperkolleg.}
\vspace{.3cm}\\
1- Ruhr West University of Applied Sciences - Dept of Computer Science \\
Lutzowstrassse $5$, $46236$ Bottrop - Germany\\
}
\maketitle
\begin{abstract}
Meta AI has recently released the Segment Anything Model (SAM), which demonstrates exceptional zero-shot image segmentation performance across various tasks with remarkable accuracy. Despite its inability to provide accurate segmentation across multiple research fields, SAM still serves as a valuable starting point for supporting the segmentation pipeline process, particularly for tasks that require extensive and senior skills annotations. This study aims to evaluate the generalization of SAM and fine-tuning SAM models using three \textbf{waste} segmentation datasets. Although they are captured from real scenes as SAM was pretrained on, these datasets present several challenges, including occlusions, deformable objects, transparency, and objects easily confused with backgrounds. In our findings, the fine-tuned SAM-ViT-H model outperforms the state-of-the-art Zerowaste, and TACO datasets with a significant increase of $+30$ in IoU, and it closely approaches performance levels of TrashCan $1.0$, with only a $-1.44$ difference. After evaluating these popular \textbf{waste} datasets, it became evident that fine-tuning SAM as a foundational model is a crucial step for providing better generalization for downstream waste segmentation tasks. Therefore, SAM should not be disregarded or \textbf{wasted}.

\end{abstract}
\section{Introduction}\label{Intro}
Deep learning models benefit from training on a lot of labeled data, like ImageNet, JFT, COCO, etc., to achieve high accuracy. So, using these benchmarking datasets to start building knowledge and then refining it for specific tasks, known as transfer learning, is essential. Large Language Models (LLM) are a recent advance, that undergo training on vast textual datasets. These models have garnered considerable attention from researchers due to their impressive performance in Natural Language Processing (NLP) tasks, like Generative Pre-trained Transformer (GPT) \cite{GPT}. GPT excels across a range of language tasks, sparking interest in real-world applications. ChatGPT has received special attention  for commercial success \cite{chatgpt}. This trend is influencing computer vision models too. Diffusion models, for instance, are reshaping text-image generation \cite{ramesh2021zeroshot}. CLIP \cite{CLIP} and ALIGN \cite{ALGN} boosted computer vision (CV), but challenges remain unresolved. These foundational CV models, while promising, are still in their early stages and currently show performance levels that are relatively less satisfactory than achievements in NLP.\\ 
Recently, Meta AI introduced SAM, a general image segmentation model. This motivates data-centric AI applications to receive substantial support from foundation models. SAM can segment ''almost" any object in any image without training (zero-shot). This capability comes from training the huge SA-$1$B dataset, which has so far the largest CV dataset, including $11$ million images and $1.1$ billion masks. The huge dataset contains images taken by photographers with their cameras, and it preserves object privacy through blurring faces and licensing plates in all images \cite{MetaSA1B}.\\
In addition, SAM represents a versatile image segmentation model that generates segmentation results for foreground, background, text, etc. It operates in three distinct modes: \textit{click mode}, where users can select objects through one or more clicks; \textit{box mode}, in which users delineate bounding boxes around objects; and \textit{everything mode}, which automatically masks all objects within an image.  SAM consists of three components: a flexible prompt encoder that can be sparse (points, boxes, text) or dense (masks), a fast mask decoder, and an image encoder based on the vision transformer model, and it is built on the vision transformer (ViT) model. SAM offers three versions with different backbone sizes: ViT-B, ViT-L, and ViT-H, with $91$M, $308$M, and $636$M parameters, respectively \cite{SAM}. 
This study explores how well SAM performs in real waste segmentation images. Although they are natural images, they have many challenges like being deformable, occluded, transparent, and easy to confuse with background objects.

\section{Challenges and potentials of SAM} \label{related}
SAM applications are vast, and some of them even extend further for ''anything tasks'', to name but a few; Inpaint Anything \cite{yu2023inpaint}, anything-$3$D \cite{shen2023anything3d}, any-to-any style transfer \cite{liu2023anytoany}, and track Anything \cite{yang2023track}. More interesting applications, e.g. \cite{zhang2023text2seg} SAM combined with Grounding DINO and CLIP for remote sensing semantic segmentation tasks via text prompts. Another application for generating pseudo labels as a prior step to pre-train thermal infrared image segmentation tasks using SAM as a teacher in knowledge distillation approach \cite{chen2023learning}\\
Although SAM proves its versatility in many applications, it has many shortcomings. For example, SAM shows successful segmentation performance in agriculture applications, such as pest and leaf disease identification. However, it requires additional knowledge of crop segmentation for example because it handles crops as a background. In addition, SAM has a relatively appealing segmentation performance on normal-sized objects, but it shows some conflicts in irregular size and shaped objects, even for blurred boundaries or small disconnected components. This applies to aerial images, remote sensing, etc. that contain small-sized objects like streets, buildings, agricultural objects, etc. \cite{tang2023sam}\\
Medical image scenarios took the lion's share of SAM applications since the segmentation task is fundamental in medical imaging. In addition, they have privacy concerns, and labeling them by clinical experts is expensive. Besides, SAM was not trained on similar images, therefore zero-shot is desired. For example, a skin lesion segmentation task faces multiple challenges since objects have vision obstacles in colors, contrast, and blurred boundaries \cite{zhou2023sam}. \\
SAM exhibits a strong generalization on natural scenes and has been tested in a broad range of experiments. While for task-specific image segmentation, SAM has limited generalization and requires additional knowledge to achieve high performance \cite{he2023accuracy}. Even for natural images, such as scenes with shadows, SAM failed to extract these shadowed objects, because it was trained on foreground objects rather than background objects. Moreover, SAM struggles with segmenting objects with similar enclosed objects like camouflaged objects with similar foreground/background image contrast or cluttered objects. Furthermore, SAM struggles to capture objects tolerant of light conditions. This includes transparent objects that reflect light like glass, plastic, bottles, etc. \cite{ji2023sam}\\
Most of the previous tasks lack training instances in task-specific domains, which requires additional examples, and fine-tuning SAM could potentially deliver better performance than zero-shot transfer more effectively. On the other hand, SAM-Adaptor is proposed instead of fine-tuning SAM model \cite{chen2023sam}. It found that SAM-Adapter enhances SAM performance and even achieves state-of-the-art for the applications specific to segmenting foreground objects, like shadow and camouflaged object detection, but it only supports single-class segmentation. According to \cite{ji2023sam}, to alleviate the open-set problem, prior knowledge is required to gain high accuracy. Therefore, there is a growing need to fine-tune the SAM model to downstream datasets to improve its robustness and generalizability.\\
\section{Method and discussion}
We chose the waste segmentation task as mentioned in section \ref{Intro}. We pretrained SAM models using three waste segmentation datasets with COCO format. This include varying datasets: For the Zerowaste dataset, we chose the full polygon annotated version that consists of the largest unbalanced deformable objects (cardboard, soft plastic, rigid plastic, and metal) with $4,661$ frames in an extremely cluttered environment that has been captured from conveyor belt \cite{zerowaste}. TrashCan $1.0$ dataset consists of $7,212$ images of underwater trash, plants, and animals. It was created for the marine robots community so they can develop efficient and autonomous trash segmentation for removal \cite{trashcan}. TACO dataset for waste in the wild consists of $1,500$ images and $4,784$ annotations in multiple environments such as woods, roads, and beaches \cite{taco}.\\
We fine-tune the mask decoder only and keep the image and prompt encoders because the original SAM uses masks, points, and bounding boxes (bboxes). However, we validated only the bbox prompts derived from the ground truth masks for ease of use \cite{Medeiros}. Practically, we used pretrained SAM weights (for ViT-L, ViT-B, and ViT-H) to initialize the weight of the mask decoder and fine-tune it in the training loop. The loss is calculated as follows: $20$ * focal loss + dice loss + MSE loss.\\
 For the tested datasets, we maintained the original dataset's training/testing split. If the split was not provided, we used an $80$\%/$20$\% ratio. We implemented the training process with the following hyperparameters: Batch-size=$8$, learning rate= $8 .10^{-4}$, with Adam optimizer on a single NVIDIA Quadro RTX $8000$ GPU with $48$ GB. We controlled overfitting using the early stop scenario when further training would not yield improvement.
The performance of mask quality was evaluated using mean Intersection over Union (IoU), which computes the overlap between ground truth and predicted masks and divides the intersection of the masks by their union. Additionally, mean precision and mean F1 pixel accuracy were calculated to determine the percentage of correctly classified pixels, as shown in Table \ref{table1}.

\begin{table}[b!]
            \footnotesize
                \begin{tabular}[t]{|p{4cm}|c|c|c|c|}
                    \hline
                   \emph{ZeroWaste Dataset (5 Epochs)} & Mean IoU & Mean Precision & Mean F$1$ Pixel Accuracy\\ \hline
                    DeepLabv3+ \cite{zerowaste}& 52.13 & 75.63 & 59.46 \\ \hline
                    SAM-ViT-B & 55.39 & 79.63 & 64.65\\ \hline
                    Fine-tuned SAM-ViT-B&79.22&87.88&86.11 \\ \hline   
                    SAM-ViT-L&57.22&80.13&65.46\\ \hline
                    Fine-tuned SAM-ViT-L&80.64&89.93&88.82 \\ \hline  
                    SAM-ViT-H&61.12&83.39&69.10 \\ \hline
                   \textbf{Fine-tuned SAM-ViT-H} &\textbf{82.79}&\textbf{85.14}&\textbf{90.53} \\ \hline
                \end{tabular}
                 \hfill \hfill \hfill \hfill
                \begin{tabular}[t]{|p{4cm}|c|c|c|c|}
                    \hline
                    \emph{TrashCan1.0 (5 Epochs)}  & Mean IoU & Mean Precision & Mean  F$1$ Pixel Accuracy\\ \hline
                    DeepLabv3+& \textbf{80.32} & \textbf{86.7} &  \textbf{91.25}\\ \hline
                    SAM-ViT-B & 62.87 & 71.2 & 74.3 \\ \hline
                    Fine-tuned SAM-ViT-B& 76.91 & 80.81&  85.04\\ \hline   
                    SAM-ViT-L& 63.38 & 71.7& 75.98 \\ \hline
                    Fine-tuned SAM-ViT-L& 77.26 & 81.21 &  86.56 \\ \hline  
                    SAM-ViT-H&66.43 & 73.30& 78.08 \\ \hline
                   \textbf{Fine-tuned SAM-ViT-H} & 78.88& 82.76&87.38 \\ \hline
                \end{tabular}
                \hfill \hfill \hfill \hfill
                \begin{tabular}[t]{|p{4cm}|c|c|c|c|}
                    \hline \hline 
                    \emph{TACO Dataset (10 Epochs)}  & Mean IoU & Mean Precision & Mean F$1$ Pixel Accuracy\\ \hline
                    DeepLabv3+&29.19 &44.86 & 33.21 \\ \hline
                    SAM-ViT-B&40.25 &51.90 & 53.21\\ \hline
                    Fine-tuned SAM-ViT-B& 50.40&60.41 & 63.20 \\ \hline   
                    SAM-ViT-L&41.33 & 53.06& 55.25\\ \hline
                    Fine-tuned SAM-ViT-L& 52.67& 61.85& 64.76 \\ \hline  
                    SAM-ViT-H& 46.29& 65.85& 59.40 \\ \hline
                   \textbf{Fine-tuned SAM-ViT-H} &\textbf{61.06 }&\textbf{71.59 }&\textbf{73.54 } \\ \hline
                \end{tabular}
                \caption{SAM and Fine-tuned SAM segmentation performance for the tested datasets}\label{table1}
            \end{table}

\begin{figure}[t!]
  \centering
  \includegraphics[width=0.95\textwidth, height =0.117\textheight]{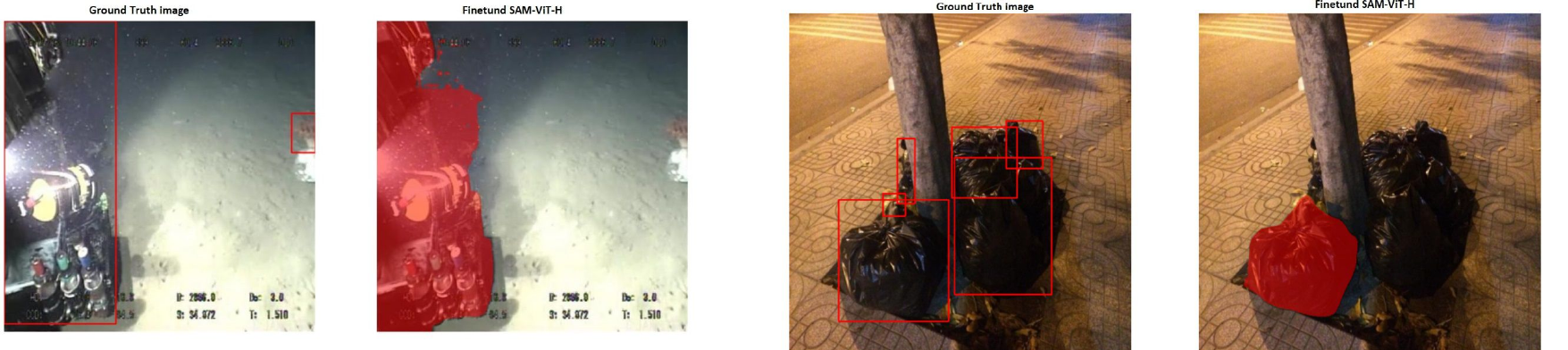}
  \caption{Segmentation results for the pair of ground truth with original box and fine-tuned SAM-ViT-H on datasets: (left) TrashCan1.0  \& (right) TACO}
  \label{fig1}
\end{figure}

As mentioned in section \ref{related}, the waste segmentation task has many challenges. The work of \cite{zerowaste} shows that popular detection and segmentation models like Mask-RCNN, TridentNet, and DeepLabV$3$+(where the latter was the best) have unsatisfactory generalization to the Zerowaste-f dataset. In addition, for this dataset in the original SAM paper \cite{SAM}, SAM has $+9.1$ mean IoU compared to RITM \cite{sofiiuk2021reviving}, which is also a strong interactive segmenter. Our fine-tuned SAM-ViT-H model demonstrates a mean IoU increase of $+30$ compared to DeepLabv$3+$, and $+21$ compared to the original SAM.\\
For the TrashCan $1.0$, a mean IoU decrease of $-15.0$ compared to RITM according to the SAM original paper \cite{SAM}, but the fine-tuned SAM-ViT-H has $-1.44$ mean IoU because SAM predictions have not matched with ground truth due to the ambiguity of the mask and small connections between objects, as shown in Figure \ref{fig1}.\\
The TACO dataset is very challenging because it was annotated roughly, and it has an extremely variable background, contrasts, outdoor variant lighting conditions, and object sizes, as shown in Figure \ref{fig1}. Therefore, SAM struggles to distinguish shadows, tiny objects, and occluded objects. 
Fine-tuned SAM-ViT-H provides the most superior performance compared to the other two versions because it has the highest number of parameters making it more general at capturing complex patterns. However, it requires more computational resources.\\
We found that SAM can be fine-tuned to create better segmentation performance for waste segmentation tasks. Therefore, fine-tuning SAM shows the ability to generalize tasks related to waste segmentation.

\section{Conclusion}
The study demonstrates the effectiveness and powerful impact of SAM as well as its limitations. We investigate the potential of leveraging SAM as a pre-trained model for segment-specific waste. Although the studied datasets were natural images, they are very challenging because of the variety of object sizes, orientation, lighting, background, and interconnection. We found that fine-tuned SAM-ViT-H is great potential since it approaches, or outperforms the state-of-the-art. The limitation of this study is that it evaluates the datasets that have been labeled based on bounding boxes. However, it can be alleviated using an additional prior step to draw them using detection models to locate them, then pass the dataset into the fine-tuned SAM. Eventually, this opens new possibilities for CV applications, like vision tasks analysis and understanding. On the other hand, it encourages the CV community to develop more robust foundation models. Future work aims to study more case studies and multiple datasets to support our findings.

\begin{footnotesize}
\bibliographystyle{unsrt}
\bibliography{ref}
\end{footnotesize}
\end{document}